\documentclass[journal]{IEEEtran}  

\IEEEoverridecommandlockouts                              

\usepackage{graphics} 
\usepackage{epsfig} 
\usepackage{mathptmx} 
\usepackage{times} 
\usepackage{amsmath} 
\usepackage{amssymb}  
\usepackage{xcolor}
\usepackage{textcomp}
\usepackage{lscape}
\usepackage[ruled,vlined]{algorithm2e}

\usepackage{subcaption}
\usepackage{lipsum}
\usepackage{cite}
\usepackage{dblfloatfix}

\makeatletter
\def\BState{\State\hskip-\ALG@thistlm}
\makeatother

\title{\LARGE \bf
Soft Soil Gait Planning and Control for Biped Robot using
Deep Deterministic Policy Gradient Approach
}

\author{Gaurav Bhardwaj$^{{1},{*}}$, Soham Dasgupta$^{2}$, N. Sukavanam$^{3}$ and  R. Balasubramanian$^{1}$
\thanks{$^{1}$Gaurav Bhardwaj and R.Balasubramanian are with the Computer Science and Engineering Department, IIT Roorkee
        {\tt\small gbhardwaj@cs.iitr.ac.in}, {\tt\small balarfcs@iitr.ac.in}}%
\thanks{$^{2}$Soham Dasgupta is with the Mechanical and Industrial Engineering Department, IIT Roorkee
        {\tt\small sohamd@me.iitr.ac.in}}%
\thanks{$^{3}$N. Sukavanam is with the Mathematics Department, IIT Roorkee
        {\tt\small nsukvfma@iitr.ac.in}}%
\thanks{$^{*}$ Corresponding Author}%
\thanks{$^{\dag}$ Funded by Council of Scientific and Industrial Research (CSIR), New Delhi under Grant No- 09/143(0903)/2017-EMR-I .}%
}

\begin{document}

\maketitle
\thispagestyle{empty}
\pagestyle{empty}


\begin{abstract}
   Biped robots have plenty of benefits over wheeled, quadruped, or hexapod robots due to their ability to behave like human beings in tough and non-flat environments. Deformable terrain is another challenge for biped robots as it has to deal with sinkage and maintain stability without falling. In this study, we are proposing a Deep Deterministic Policy Gradient (DDPG) approach for motion control of a flat-foot biped robot walking on deformable terrain. We have considered a 7-link biped robot for our proposed approach. For soft soil terrain modeling, we have considered triangular Mesh to describe its geometry, where mesh parameters determine the softness of soil. All simulations have been performed on PyChrono, which can handle soft soil environments.
 
\end{abstract}
\section{INTRODUCTION}

Humanoid robots are currently one of the widely researched topic out there.Humanoid robots or Biped robots generally have a highly complex and  nonlinear model which makes it quite difficult to establish its dynamic model\cite{r1}. There have been many analyses on the control strategy of the Biped robot. Although most of the research deals with assumptions that the robot walking on hard or uneven ground\cite{r2},\cite{r3}, very few have actually researched about walking dynamics for a bipedal robot in a deformable or soft terrain\cite{r4},\cite{r12}. Until now most of the research focused on model and terrain interaction with mostly experimental and numerical simulations with a terrain model with dealing with just static sinkage\cite{r6},\cite{r7}. 

The terrain model we implemented is based on generalization of the Bekkar-Wong model combined with multi-body dynamics simulation of Project Chrono.The main advantage of using Chrono over other simulation software is the lightweight formulation that allows real time contact force modeling\cite{r8} . The terrain model can be calculated as a generalization of the Bekker-Wong model for general three dimensional shapes\cite{r9},\cite{r10},\cite{r11},\cite{r12}.

\[ \sigma = (\frac{k_{c}}{b} + k_{phi})\*y^n \]
where \(\sigma\) denotes the contact pressure of the patch, $y$ represents the associated sinkage, \(k_{c}\)  is a coefficient representing the cohesive effect of the soil, \(k_{phi}\) represents the soil stiffness, $n$ denotes the effect of hardening and $b$ is the range of shorter side considering the rectangular contact footprint.

For a general contact footprint, the length b is little ambiguous to define ;hence instead, we estimate this length by first obtaining all connected contact patches and then using the approximation
\[ b \approx (\frac{2*A}{L}) \]
where A is the area of such a contact patch and L its perimeter.The dynamic sinkage analyis of the biped robots is performed at every timestamps of the multi-body dynamics simulation.There is already a similar case study on a hexapod robot model based on the multi-body dynamics of Project Chrono\cite{r13}

These conventional control theory approaches rely on the complex deterministic and mathematical engineering models depending on extensive mathematical computations and complete knowledge of the dynamical system of robot and terrain interactions\cite{r14},\cite{r15}.Reinforcement learning practices in the recent years have proved their advantage over the conventional classical control system problems and have been successful in stable bipedal walking \cite{r16},\cite{r17}.DDPG  or Deep Deterministic Policy Gradient algorithm is a model free off policy learning method for learning continuous actions which is widely used to solve control theory problems \cite{r18}. DDPG can be directly used to control the model based upon the raw continuous inputs gathered from the simulation.It uses end to end machine learning to control the vouch in the hope that it uses mini-batch training to reduce the training time\cite{r19},\cite{r20}.

Furthermore, this paper is organized as follows. In Section 2 our biped robot model with associated parameters is elaborated. Section 3 provides a glimpse of Reinforcement Learning with focus on DDPG. Section 4 describes our contact model for deformable soil followed by simulation results and conclusions in section 5 and section 6 respectively.

 \section{Biped Robot Model and Parameters}
\begin{figure}[h]
	\centering	 
	\captionsetup{justification=centering} 
	\subfloat[Frontal Plane View\vspace{-.05in}]
	{\includegraphics[width=0.20\textwidth]{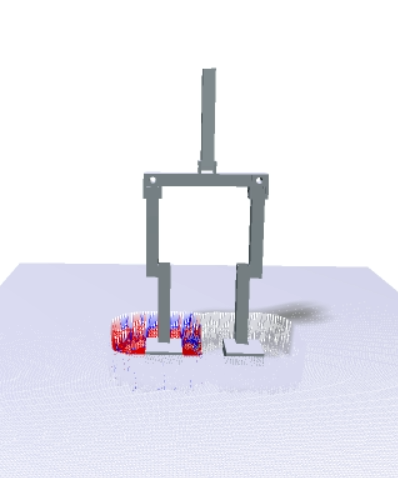}\label{fig:fpv}} 
	\hspace{.01cm}
	\subfloat[Sagittal Plane View\vspace{-.05in}]
	{\includegraphics[width=0.18\textwidth]{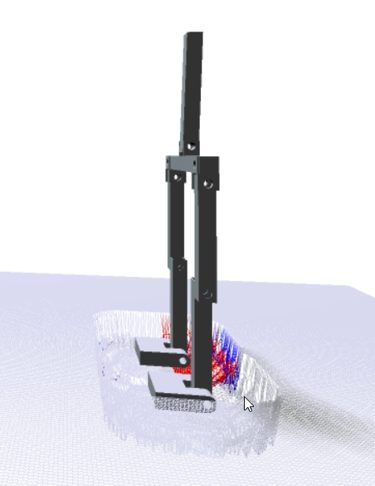}\label{fig:spv}} 
	\caption{7-Link biped Robot Model} 
	\label{fig:chronobiped}		
\end{figure}

We have developed a CAD model of a 7-Link biped robot with a flat foot using SolidWorks, as shown in figure \ref{fig:chronobiped} and imported it into the PyChrono simulation environment with frontal (figure \ref{fig:fpv}) and sagittal plane (figure \ref{fig:spv}) views. It has seven degrees of freedom; one torso, one hip $\times 2$, one knee $\times 2$, and one ankle  $\times 2$. 
Table \ref{tab:pcrcm} shows parameters for the considered biped robot.
\begin{table}[]
    \centering
    \begin{tabular}{|c|c|c|}
    \hline
       Links  & Mass (Kg) & Length (mm) \\
       \hline
        Thigh Link &  2.5 & 230 \\
        Lower Leg Link & 2.5 & 230 \\
        Hip Joint & 1.5 & 130 \\
        Torso & 4 & 230 \\
        Foot & 1 & 90 \\
        \hline
    \end{tabular}
    \caption{Biped Robot Parameters}
    \label{tab:pcrcm}
\end{table}
\section{Reinforcement Learning}
Human beings interact with the environment and subsequently take actions to learn it, and based on the experience they get after the interaction, they decide their further actions. Let us take an example where a burning candle is put before a child who knows nothing about it. He will first try to touch the flame, and after getting hurt, he learns not to get near the candle again. This type of scenario where an agent in a particular state wants to fulfill a task has to go through a series of states by taking some actions. He will be penalized for wrong actions and will get rewarded for correct actions, and similarly, he will try to develop a policy to maximize its cumulative rewards.
Mathematically, this approach to learning is known as Reinforcement Learning.

\begin{figure}
    \centering
    \includegraphics[width=1.0\linewidth]{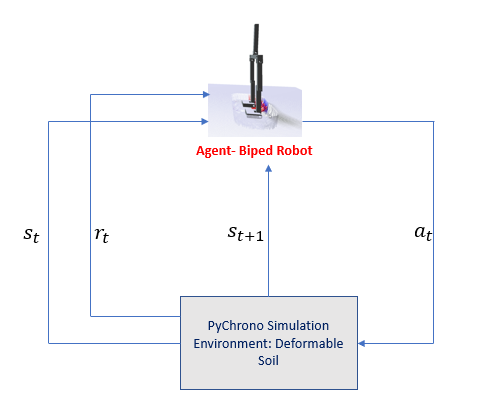}
    \caption{Biped Robot Agent Interaction in PyChrono Environment}
    \label{fig:intr}
\end{figure}

We have an agent, due to its interaction with the environment, that will get an observation space $s_t$ at a particular time instant $t$, which describes the state of the agent at that time instant. The agent also has an action space which is the set of actions from which it can choose action $a_t$ to take in a particular state $s_t$. It will receive reward $r_t$ and will move to new state $s_{t+1}$ from $s_t$ \cite{r18} as shown in figure \ref{fig:intr}.  
Policy $\pi$ maps state to action. it determines which action will be taken by an agent in a particular state. The policy can be deterministic [$\pi(S)=A$] or stochastic [$\pi:S\rightarrow{P(A)}$] where probability distribution over action space is established, where $S\ and\ A$ represents State Space and Action Space respectively. Markov Decision Process (MDP) can be considered to model the environment. Return $R_t=\sum^{t}_{i=1} \gamma^{i-t}r(s_i,a_i)$ is discounted sum of future rewards, where $\gamma$ is discount factor that reduces the weightage of future rewards due to discrepancies related to future predictions. The agent's ultimate goal is to maximize the expected return \cite{r19}.  Q-Function or action-value function  accounts for the expected return as described below:
\begin{equation}
    Q^{\pi}(s_t,a_t)= \mathbb{E}_{r_i>=t,s_i>t \sim E, a_i>t \sim \pi }[R_t|s_t,a_t]
\end{equation}
For deterministic policy $\mu$, the equation can be rewritten as follows:
\begin{equation}
    Q^{\pi}(s_t,a_t)= \mathbb{E}_{r_t,s_{t+1} \sim E}[r(s_t,a_t)+\gamma Q(s_{t+1},\mu (s_{t+1}))]
\end{equation}
Q-Leaning is an off-policy algorithm with a greedy approach-based policy $\mu(s)=argmax_aQ(s,a)$.
\subsection{DDPG}
DDPG is an powerful off-policy model free learning method for a continuous action space. It can be considered as similar to Deep Q Learning (DQN) except for the fact it learns over an continuous action space over a discrete action space. It uses off-policy data to learn Q-Value function and uses Q- Value function to learn the optimal policy. 
One of the general features of Q-Learning networks is that they use target networks.  The DDPG uses four neural networks the current Q-function \(Q(s,a|\theta^Q)\), the policy \(\mu(s|\theta^\mu) \), the targeted Q-function \(Q^{'}(s,a|\theta^{Q^{'}})\)  and the targeted policy \(\mu^{'}(s|\theta^{\mu^{'}})\)  , where \(\theta^Q\), \(\theta^\mu\), \(\theta^{Q^{'}}\) and \(\theta^{\mu^{'}}\)  are the weights of each networks. The target networks are just copies of the original networks with a time delay which is sampled from the original networks after every fixed number of iteration of the networks.
The updated Q-Value function using target functions and bellman equations is obtained as follows :
\[y_{i} = r_{i}+\gamma Q_{i}(s_{i+1},\mu^{'}(s_{i+1}|\theta^{\mu^{'}})|\theta^{Q^{'}})\]
The loss functions is the squared mean error between the updated Q-Value function and the original one.Hence the Q-Value function is updated at each step by minimising the loss function:
\[ L(\theta^{Q}) = \frac{1}{N}\sum_{\mu }^{}[(y_{t}-Q(s_{t}-Q(s_{t},a_{t}|\theta ^{Q}))^2)]\]
The policy is obtained using batch gradient descent using the function:
\[\nabla_{\theta^{\mu }}J(\theta ) \approx \frac{1}{N}\sum_{i}^{}[\nabla_{a}Q(s,a|\theta^Q)|_{s=s_{t},a=\theta(s_{t})}\triangledown_{\theta^{\mu }}\mu (s|\theta^\mu)|_{s=s_{t}}] \]
The updates for the target networks for both policy and Q-function are given by
\[\theta^{Q^{'}}\leftarrow \tau \theta^{Q}+(1-\tau)\theta^{Q^{'}}\]
\[\theta^{\mu^{'}}\leftarrow \tau \theta^{\mu}+(1-\tau)\theta^{\mu^{'}}\]
DDPG also makes use of replay buffer which is essential for stable networks . It stores information about recent experiences to update the hyper-parameters.

\section{Deformable Soil Contact Model}
\begin{figure}
    \centering
    \includegraphics{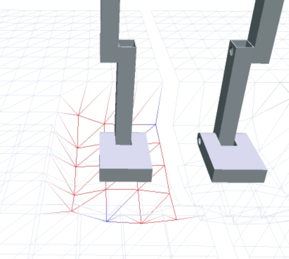}
    \caption{Deformable Soil Contact Model}
    \label{fig:mesh}
\end{figure}
The soil-contact model is defined using Deformable SCM (Soil-Contact Model) from the project Chrono library. The surface of the terrain is a Cartesian rectangular mesh with Nodes (vertices) that serve as the contact point of the soil terrain interaction. The forces are calculated from the vertical displacement of the nodes resulting from the foot-terrain interaction. The deflection of the nodes is the only data that is stored in the hash maps, which makes the processing quite fast compared to other methods.
The normal force is calculated using Bekkar -Wong model. The pressure on each node(vertices) is given by
\[\sigma _{node} = K_{\phi }*Y_{node}^{n}+R*V_{node}\]
where \(\sigma_{node}\) is the Bekkar friction modulus, \(Y_{node}\) is the displacement of the node, n is the sinkage exponent, R is the damping coefficient, and \(V_{node}\) is the normal velocity of the node.
The tangential forces experienced by the robot is based on the Coulomb-Mohr theory criteria. The maximum shear that can be experienced by the node is given by 
\[\tau_{max } = \sigma_{normal}*tan\theta + c\]
where c is the Mohr Cohesion. , \(\theta\) is the friction angle , \(\sigma_{normal}\) is the normal stress from the node and \(\tau_{max}\) is the maximum shear stress from the particular node.
Due to slipping, the robot won't experience the maximum shear stress all the time. Hence the tangential shear stress with slip condition is  calculated  from the Janoshi-Hanamoto equation of shear stress which is given by :
\[\tau_{node} = \tau_{max}(1-\exp(-\frac{b_{node}}{K}))\]
where \(\tau_{node} \) is the shear stress experienced by the node, \(\tau_{max}\) is the maximum shear stress as calculated above, \(b_{node}\) is the shear displacement of the node and K is the shear parameters.
The resultant force on the robot feet is calculated from the summation of all the contact forces from all the vertices in contact with the robot feet. The robot parameters can be customized with the experimental approach. For our case, we went with the following parameters. The soil contact parameters are given in Table \ref{tab:scmp}.
\begin{table}[]
    \centering
    \begin{tabular}{|c|c|c|}
    \hline
       Soil Parameters  & Value\\
       \hline
        Bekker \(k_{\phi}\)  &  0.2e6  \\
        Bekker \(k_{c}\) & 0  \\
        Bekker n & 1.1  \\
        Mohr cohesion & 0  \\
        Mohr Friction & 30  \\
        Janoshi Shear & 0.01  \\
        Elastic Coefficient & 4e7  \\
        Damping Coefficient & 3e4  \\
        Friction Angle & 30 \\
        \hline
    \end{tabular}
    \caption{Soil Contact Model Parameters}
    \label{tab:scmp}
\end{table}

\section{Simulation Results}
\begin{figure*}[htbp]
\begin{subfigure}{.242\textwidth}
  \centering
  \includegraphics[width=1.25\linewidth]{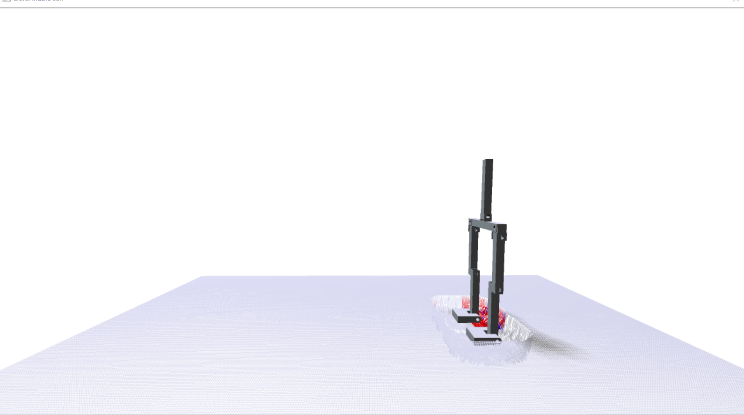}
\end{subfigure}%
\begin{subfigure}{.242\textwidth}
  \centering
  \includegraphics[width=1.25\linewidth]{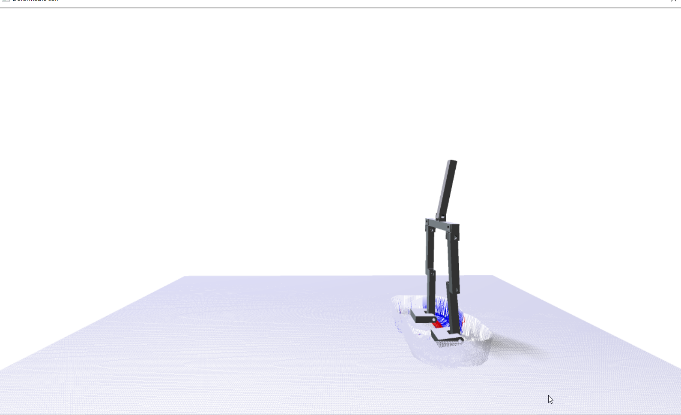}
\end{subfigure}
\begin{subfigure}{.242\textwidth}
  \centering
  \includegraphics[width=1.25\linewidth]{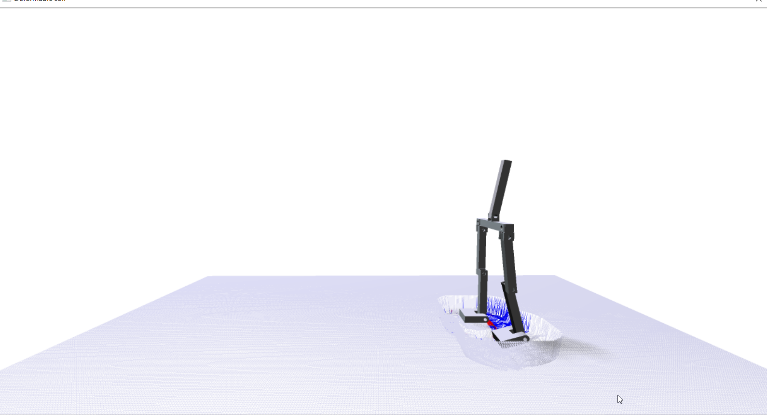}
\end{subfigure}
\begin{subfigure}{.242\textwidth}
  \centering
  \includegraphics[width=1.25\linewidth]{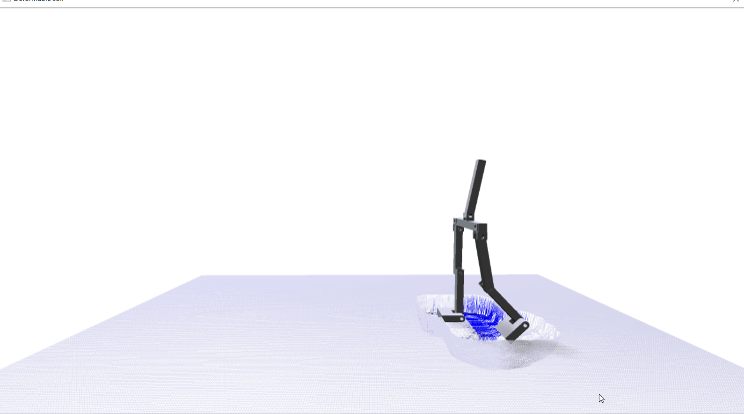}
\end{subfigure}
\begin{subfigure}{.242\textwidth}
  \centering
  \includegraphics[width=1.25\linewidth]{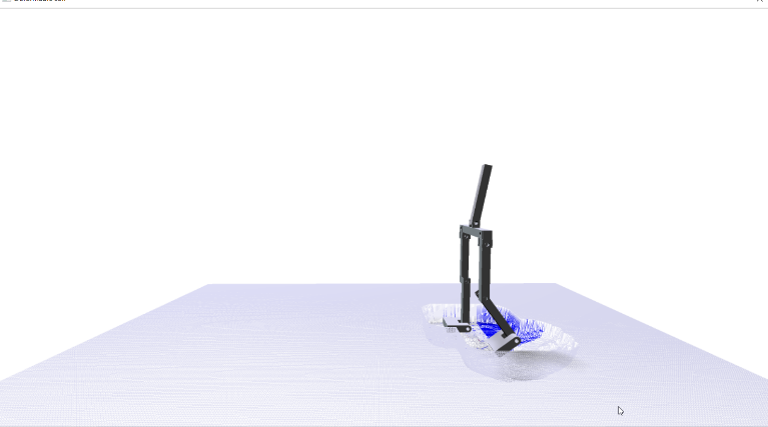}
\end{subfigure}%
\begin{subfigure}{.242\textwidth}
  \centering
  \includegraphics[width=1.25\linewidth]{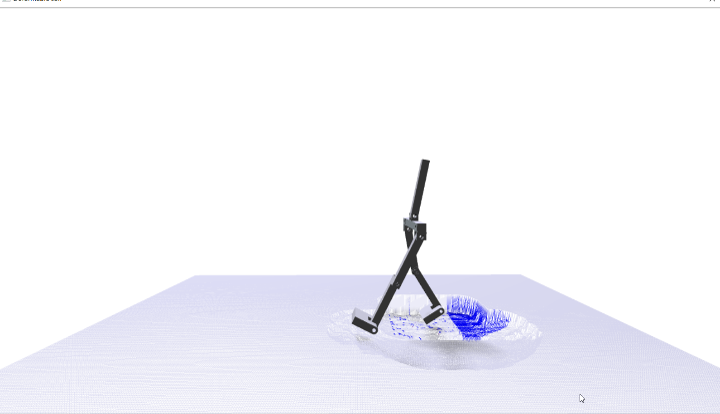}
\end{subfigure}
\begin{subfigure}{.242\textwidth}
  \centering
  \includegraphics[width=1.25\linewidth]{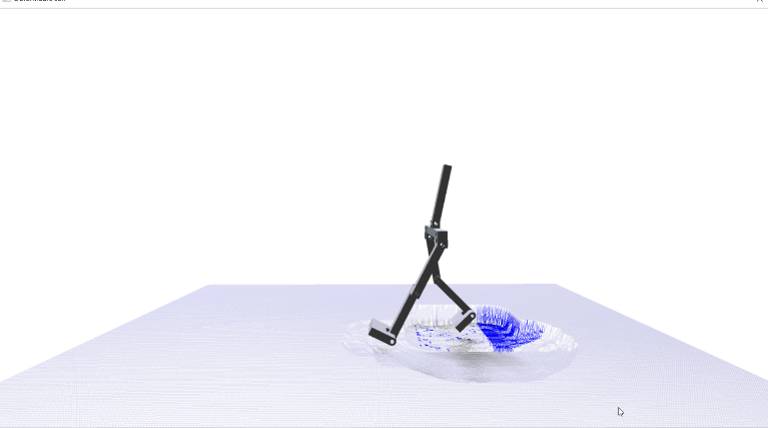}
\end{subfigure}
\begin{subfigure}{.242\textwidth}
  \centering
  \includegraphics[width=1.25\linewidth]{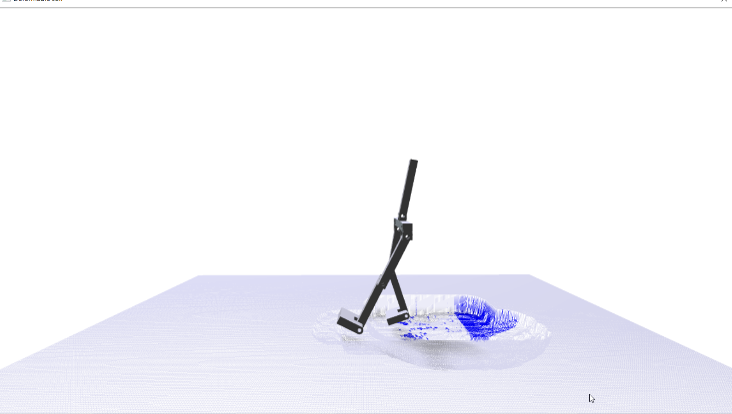}
\end{subfigure}
\caption{Snapshots of obtained simulation result}
\label{fig:introsmr}
\end{figure*}

This section is dedicated to the obtained simulation results with the PyChrono environment. The biped robot was trained and simulated on NVIDIA GeForce GTX 1050 Ti Graphical Processing Unit(GPU). It took almost 53 hours to train the model to get a stable walk in a deformable soil environment.

For our model, straight standing with a flat foot is a neutral zero radian position. The discrete Element Method is used for contact modeling in PyChrono environment. The cycloidal trajectory is considered for the ankle as a reference trajectory with inverse kinematics solution using PyChrono simulation software strategy. The goal of the agent here is to move in a straight line. Observation space for the agent consists of: Center of Mass lateral and vertical translation, angular positions \& angular velocities of all joints, forward, lateral \& vertical translation velocities, and action values from the previous time step. Action space consists of joint space, which the agent can control by applying torques. The episode terminates when our agent or biped robot falls down. The agent is rewarded for forward translation and penalized for lateral \& vertical translations. Figure \ref{fig:introsmr} shows sequential instances of simulation results. The agent fell down after a 10 m walk.
The simulation results provided us with the contact forces acting on the feet from the terrain.

\begin{figure}[htbp]
    \begin{subfigure}{.242\textwidth}
        \centering
        \includegraphics[width=1.25\linewidth]{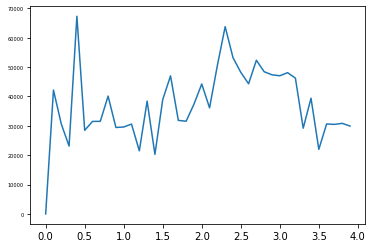}
        \caption{Normal Force on the right foot}
        \label{fig:mesh}
    \end{subfigure}
    
    \begin{subfigure}{.242\textwidth}
        \centering
        \includegraphics[width=1.25\linewidth]{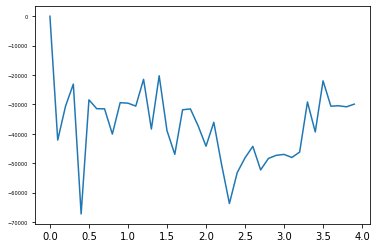}
        \caption{Normal force on the left foot}
        \label{fig:mesh}
    \end{subfigure}
    
    
\end{figure}

\section{CONCLUSIONS}

In this paper, we have analyzed biped robot walking on deformable soil, which is required in agricultural applications, harsh environments, planetary exploration, etc. We have considered PyChrono simulation software which provides a realistic deformable soil environment. A DDPG-based control approach is proposed, which showed quite promising results for our biped robot model. In future we will try to increase Degrees of Freedom of our biped robot with frontal plane motion as well.



\begin{thebibliography}{99}







\bibitem{r17} Lucian Bus¸oniu, Tim de Bruin, Domagoj Toli´c, Jens Kober, and Ivana
Palunko. Reinforcement learning for control: Performance, stability, and
deep approximators. Annual Reviews in Control, 46:8–28, 2018.
\bibitem{r1} Steve Collins, Andy Ruina, Russ Tedrake, and Martijn Wisse. Efficient
bipedal robots based on passive-dynamic walkers. Science (New York,
N.Y.), 307:1082–5, 03 2005.
\bibitem{r7} Liang Ding, Haibo Gao, Zongquan Deng, Yuankai Li, and Guangjun
Liu. New perspective on characterizing pressure–sinkage relationship
of terrains for estimating interaction mechanics. Journal of Terramechanics,
52:57–76, 2014.
\bibitem{r4} Kenji Hashimoto, Hyun-jin Kang, Masashi Nakamura, Egidio Falotico,
Hun-ok Lim, Atsuo Takanishi, Cecilia Laschi, Paolo Dario, and Alain
Berthoz. Realization of biped walking on soft ground with stabilization
control based on gait analysis. In 2012 IEEE/RSJ International
Conference on Intelligent Robots and Systems, pages 2064–2069. IEEE,
2012.
\bibitem{r13} Gang He, Zhaoyuan Cao, Qian Li, Denglin Zhu, and Ji Aimin. Influence
of hexapod robot foot shape on sinking considering multibody dynamics.
Journal of Mechanical Science and Technology, 34(9):3823–3831, 2020.
\bibitem{r9} Rishad A Irani, Robert J Bauer, and Andrew Warkentin. Application of
a dynamic pressure-sinkage relationship for lightweight mobile robots.
International Journal of Vehicle Autonomous Systems, 12(1):1–23, 2014.
\bibitem{r3} Shishir Kolathaya and Aaron D Ames. Achieving bipedal locomotion
on rough terrain through human-inspired control. In 2012 IEEE
international symposium on safety, security, and rescue robotics (SSRR),
pages 1–6. IEEE, 2012.
\bibitem{r6} Shunsuke Komizunai, Atsushi Konno, Satoko Abiko, and Masaru
Uchiyama. Development of a static sinkage model for a biped robot
on loose soil. In 2010 IEEE/SICE International Symposium on System
Integration, pages 61–66. IEEE, 2010.
\bibitem{r12} Masahiro Komuta, Yoshitaka Abe, and Seiichiro Katsura. Walking control
of bipedal robot on soft ground considering ground reaction force. In
2017 IEEE/SICE International Symposium on System Integration (SII),
pages 318–323. IEEE, 2017.

\bibitem{r19} Arun Kumar, Navneet Paul, and SN Omkar. Bipedal walking robot using
deep deterministic policy gradient. arXiv preprint arXiv:1807.05924,
2018.
\bibitem{r14} Ryo Kurazume, Shuntaro Tanaka, Masahiro Yamashita, Tsutomu
Hasegawa, and Kyushu Yoneda. Straight legged walking of a biped
robot. In 2005 IEEE/RSJ International Conference on Intelligent Robots
and Systems, pages 337–343. IEEE, 2005.
\bibitem{r18} Timothy Paul Lillicrap, Jonathan James Hunt, Alexander Pritzel, Nicolas
Manfred Otto Heess, Tom Erez, Yuval Tassa, David Silver, and
Daniel Pieter Wierstra. Continuous control with deep reinforcement
learning, September 15 2020. US Patent 10,776,692.
\bibitem{r11} Cong Ma, Fan Yu, and Zhe Luo. Simulations and experimental research
on a novel soft-terrain hexapod robot. Int. J. Robot. Autom, 30:247–255,
2015.
\bibitem{r15} Volodymyr Mnih, Koray Kavukcuoglu, David Silver, Andrei A Rusu,
Joel Veness, Marc G Bellemare, Alex Graves, Martin Riedmiller, Andreas
K Fidjeland, Georg Ostrovski, et al. Human-level control through
deep reinforcement learning. nature, 518(7540):529–533, 2015.
\bibitem{r16} Richard S Sutton and Andrew G Barto. Reinforcement learning: An
introduction. MIT press, 2018.
\bibitem{r8} Alessandro Tasora, Radu Serban, Hammad Mazhar, Arman Pazouki,
Daniel Melanz, Jonathan Fleischmann, Michael Taylor, Hiroyuki
Sugiyama, and Dan Negrut. Chrono: Multi-physics simulation engine.
Astrophysics Source Code Library, pages ascl–2009, 2020.
\bibitem{r20} Xiaoguang Wu, Shaowei Liu, Tianci Zhang, Lei Yang, Yanhui Li, and
Tingjin Wang. Motion control for biped robot via ddpg-based deep
reinforcement learning. In 2018 WRC Symposium on Advanced Robotics
and Automation (WRC SARA), pages 40–45. IEEE, 2018.
\bibitem{r2} Cong Yan, Fumihiko Asano, Yanqiu Zheng, and Longchuan Li. Analysis
of biped robot on uneven terrain based on feed-forward control. In
Climbing and Walking Robots Conference, pages 37–39. Springer, 2021.
\bibitem{r10} Huaiguang Yang, Liang Ding, Haibo Gao, et al. Experimental study and
modeling of wheel’s steering sinkage for planetary exploration rovers.
Journal of Mechanical Engineering, 53(8):99–108, 2017.

\end{thebibliography}
\end{document}